\pdfoutput=1
\documentclass[11pt]{article}
\usepackage[]{ACL2023}

\usepackage{times}
\usepackage{latexsym}
\usepackage[T1]{fontenc}
\usepackage[utf8]{inputenc}
\usepackage{microtype}
\usepackage{inconsolata}
\usepackage{xcolor}

\usepackage{color, colortbl}
\usepackage{booktabs}
\usepackage[algo2e,ruled,vlined]{algorithm2e} 
\usepackage{amsmath,amsfonts,bbm}
\usepackage{multirow}
\usepackage{booktabs,adjustbox}
\usepackage{svg} % for figures
\definecolor{prompt}{RGB}{255,228,196}

\title{Instruction Fusion: Advancing Prompt Evolution through Hybridization}

\author{Weidong Guo \thanks{\ \ These authors contributed equally to this work.} $^1$\ 
Jiuding Yang \footnotemark[1] $^2$\ 
Kaitong Yang \footnotemark[1] $^1$\  
Xiangyang Li $^1$ \\
\textbf{Zhuwei Rao} $^1$\ 
\textbf{Yu Xu} $^1$\ 
\textbf{Di Niu} $^2$\  \\
\text{$^1$Platform and Content Group, Tencent}\\
\text{$^2$University of Alberta}\\
\texttt{$^1$\{weidongguo,kaitongyang,xiangyangli,evanyiu,henrysxu\}@tencent.com}\\
\texttt{$^2$\{jiuding,dniu\}@ualberta.ca}
}

\begin{document}
\maketitle

\begin{abstract}
The fine-tuning of Large Language Models (LLMs) specialized in code generation has seen notable advancements through the use of open-domain coding queries. Despite the successes, existing methodologies like \textit{Evol-Instruct} encounter performance limitations, impeding further enhancements in code generation tasks. This paper examines the constraints of existing prompt evolution techniques and  introduces a novel approach, \textit{Instruction Fusion} (IF). IF innovatively combines two distinct prompts through a hybridization process, thereby enhancing the evolution of training prompts for code LLMs. Our experimental results reveal that the proposed novel method effectively addresses the shortcomings of prior methods, significantly improving the performance of Code LLMs across five code generation benchmarks, namely HumanEval, HumanEval+, MBPP, MBPP+ and MultiPL-E, which underscore the effectiveness of \textit{Instruction Fusion} in advancing the capabilities of LLMs in code generation.
\end{abstract}

\section{Introduction}
\label{sec:intro}

The field of automatic program writing has intrigued computer scientists since the 1960s \cite{waldinger1969prow, dehaerne2022code}. This period has seen substantial efforts to enable machines to autonomously write correct programs. The emergence of Large Language Models (LLMs) \cite{GPT3, ouyang2022training, llama} has been a cornerstone in text generation \cite{Liu_2023}, leading to the evolution of Code Large Language Models (Code LLMs) which have notably advanced code generation tasks \cite{hou2023large}.

Early Code LLM research \cite{codellama, starcoder} concentrated on the pre-training phase, utilizing a variety of code datasets to bolster the code generation capabilities of LLMs. A major breakthrough was achieved with instruction tuning \cite{wei2022finetuned}, which enhanced the general applicability of LLMs by fine-tuning them with instructions rather than task-specific prompts.

In pursuit of greater quantity, diversity, and creativity in instruction-response samples, \citet{self-instruct} introduced the \textsc{Self-Instruct} method, leveraging LLMs to generate their own synthetic instructions. This approach was notably adopted by Code Alpaca \cite{codealpaca}, based on Stanford Alpaca \cite{alpaca}, transforming 21 basic code prompts into 20,000 high-quality instructions. Building on this, \citet{wizardcoder} further advanced code generation performance by employing \textit{Evol-Instruct} on these instructions, expanding them to 78,000 through five dataset evolution heuristics.

However, the \textit{Evol-Instruct} method has its limitations on Code LLMs, primarily its reliance on a set of five heuristics for generating new instructions. This approach leads to a pattern where the evolution is done primarily by adding more constraints to its seed instruction, which can cause three major issues. Firstly, the evolving instructions may become excessively complex, challenging GPT's ability to respond effectively. Secondly, the newly added constrains to the seed instruction may not exist in the original seed instruction, leading to an gap on difficulty gradient. Finally, the evolutionary evaluations often remain confined within the scope of the original instruction, limiting diversity.

\begin{figure*}[h]
\centering
\includegraphics[width=1\textwidth]{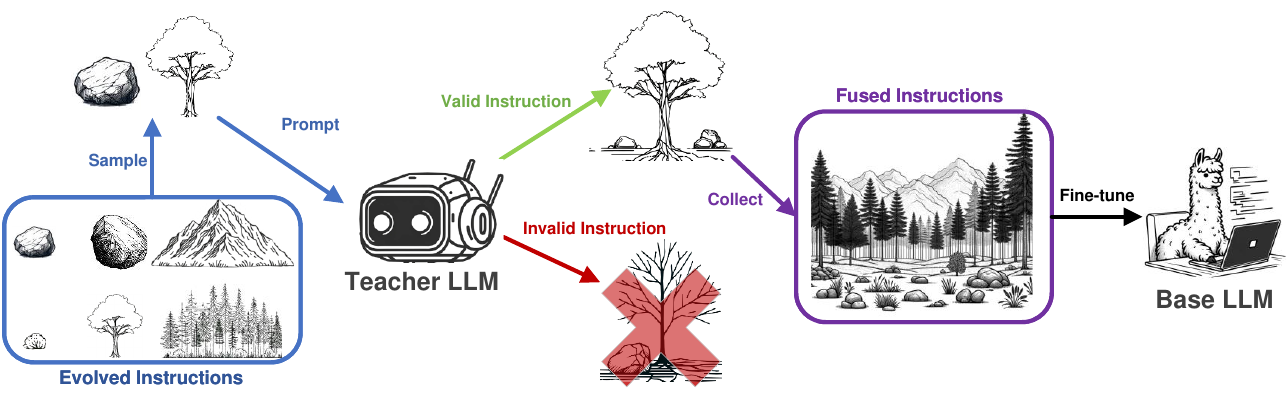} % Reduce the figure size so that it is slightly narrower than the column.
\caption{An illustration of \textit{Instruction Fusion}. The responses to each fused instruction are also generated and collected from GPT-4 Turbo.}
\label{fig-illu}
% \vspace{-5mm}
\end{figure*}

Addressing these challenges, we introduce a novel method named \textit{Instruction Fusion} (IF). This method leverages hybridization concepts to significantly improve prompt evolution, as demonstrated in Figure~\ref{fig-illu}. It involves merging two distinct instructions into a single prompt using GPT-4 Turbo, enhancing prompt complexity and diversity. This approach also facilitates a more gradual increase in difficulty for LLMs, optimizing learning and performance.

To assess the efficacy of our \textit{Instruction Fusion} method, we conducted experiments with \textsc{Code-Llama} \cite{codellama}, using HumanEval \cite{humaneval}, MBPP \cite{mbpp}, HumanEval$^+$, MBPP$^+$ \cite{evalplus}, and MultiPL-E \cite{multi} as benchmark datasets. 
Our results demonstrate a significant performance improvement of LLMs when training with the additional data generated by IF.
The contributions of this work can be summarized as follows:
\begin{itemize}
    \item {The IF method greatly improved instruction creation by merging two distinct instructions into a single, more complex prompt using GPT-4 Turbo. This strategy substantially boosts the diversity and complexity of the training data while ensuring a smoother difficulty gradient for learning. This approach effectively addresses and overcomes the limitations inherent in previous methods such as \textit{Evol-Instruct}.}

    \item{Through extensive fine-tuning experiments on Code-Llama models on five commonly used code generation benchmarks, we demonstrate the superior performance of our \textit{Instruction Fusion} method. We fully open source the model weights, training data, and source code to facilitate future research.}
\end{itemize}

% Since the evolved instruction dataset from WizardCoder is not publicly accessible, we used the \texttt{evol-codealpaca-v1} dataset, recreated by an independent party using similar techniques to WizardCoder. 
% This provided a comparable basis to evaluate the impact of our \textit{Instruction Fusion} method on code generation tasks. 
\section{Approach}
\label{sec:method}
In this section, we first analyze the the limitation of current evolution method on Code LLMs, then we elaborate on the details of the proposed \textit{Instruction Fusion} method and analyze its advantages comparing to the existing instruction evolution method. The main process of the fusion is illustrated in Figure~\ref{fig-illu}.

\subsection{Instruction Evolution}
High-quality instructions are vital for effectively fine-tuning Code LLMs. Crafting these, particularly for coding tasks, is resource-intensive and often results in easy difficulty levels, creating a gap in challenging content \cite{wizardLM}. To address this and reduce costs, \citet{wizardcoder} applied \textit{Evol-Instruct} \cite{wizardLM} to Code LLMs, enhancing code generation. This process involves merging each seed instruction with a unique evolution prompt and generating evolved instructions using GPT-3.5 Turbo. This method expanded and diversified the instruction dataset, but certain limitations hinder its further development.

\begin{figure}[t]
\centering
\includegraphics[width=1\columnwidth]{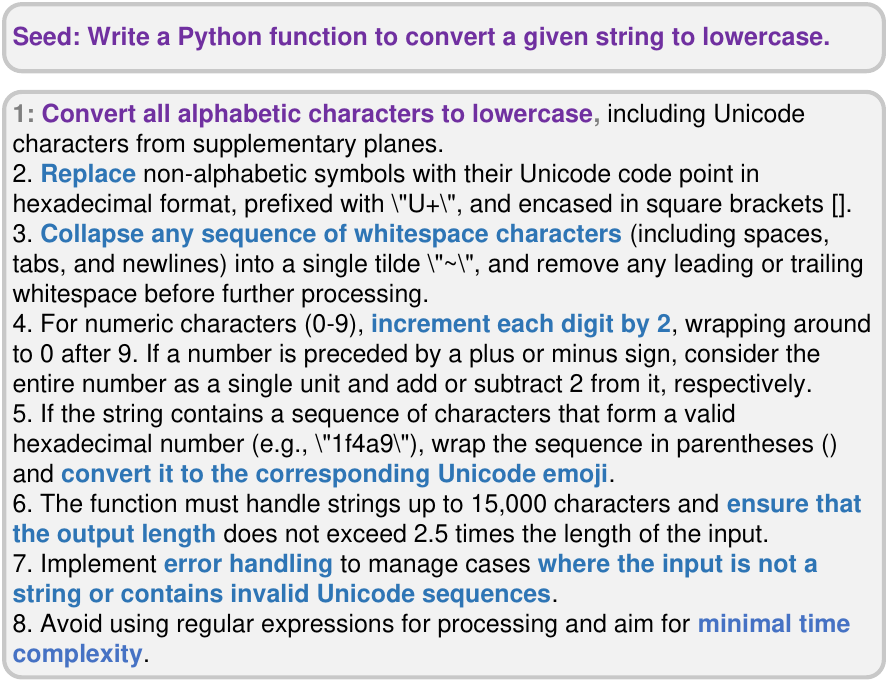} % Reduce the figure size so that it is slightly narrower than the column.
\caption{A real example of the fourth round evolution using Evol-Instruct.}
\label{fig-case-lower}
% \vspace{-5mm}
\end{figure}

A major challenge identified in the process of instruction evolution is the over-escalating constraints, as illustrated in Figure~\ref{fig-case-lower}. The process begins with a straightforward instruction, such as developing a Python function to convert strings to lowercase. However, the complexity increases significantly with each evolution round. After four rounds, the instruction length expanded from 14 to 325 tokens. This increase is primarily attributed to the incorporation of up to eight distinct constraints. As a result, the evolved instructions become impractically intricate and overly complex. 
% For a comprehensive statistical analysis of the instruction length in relation to the number of evolution rounds, please see the detailed data presented in the Appendix.

Although \textit{Evol-Instruct} with low rounds can properly increase the difficulty levels by braining new constrains, which benefits LLM learning, those new constrains may need prior knowledge to learn. Consider math learning as an analogy: understanding addition and multiplication is fundamental before tackling problems like "$1 + 1 \times 2$." However, \textit{Evol-Instruct} might prematurely introduce such problems immediately following basic concepts like "$1 + 1 = 2$", potentially leading to gaps in learning progression.

Another issue is the limited diversification of objectives during evolution. Despite adding new constraints, the primary objective often remains unchanged. Take the scenario in Figure~\ref{fig-case-lower} as an example,  %started with a Python print task ("Hello, World!"). 
despite introducing new objectives like sorting, the primary task remained similar to the original. This lack of variability in objectives can severely limit the diversity of the resulting instructions.

Due to these inherent limitations, the \textit{Evol-Instruct} method applied to Code LLMs often reaches its capacity (3 rounds) without further enhancing code generation performance. However, these issues can be mitigated with our proposed \textit{Instruction Fusion} method.

\subsection{Instruction Fusion}
Inspired by the hybridization which can produce new individual by exchange the information between inter-specific parents, we propose \textit{Instruction Fusion} to overcome the above challenges. 

\textbf{Fusion Process.}
Figure~\ref{fig-illu} gives an illustration of the \textit{Instruction Fusion} process.
Denote the initial dataset as $\mathbf{C}=\{(I_i,R_i)\}_{1<i<N}$, where $I_i$, $R_i$ are the $i^\text{th}$ seed instruction and the corresponding response, $N$ is the number of instructions in $\mathbf{C}$. To create a new instruction using \textit{Instruction Fusion}, we first random select two seed instructions $I_j$ and $I_k$ from $\mathbf{C}$, where $1\le j,k\le N$. Then, denote the target amount of fused instructions as $M$, we prompt GPT-4 Turbo to fuse the selected two instructions into the $m^\text{th}$ fused instruction $I^m_{(j,k)}$ 
 utilizing the following prompt:\\

\noindent\emph{Your task is to act as a Prompt Fusion Specialist. Your objective is to merge \#Given Prompt 1\# and \#Given Prompt 2\# into a single, cohesive \#Fused Prompt\#. This new prompt should:\\
1. Integrate the content from both \#Given Prompt 1\# and \#Given Prompt 2\#.\\
2. Maintain a similar length and complexity level as the original prompts.\\
3. Be coherent and solvable, incorporating elements from both prompts in a balanced way.\\
4. In cases where the original prompts specify different programming languages, choose only one for the \#Fused Prompt\#.\\
If the resulting \#Fused Prompt\# is not logically coherent or solvable, simply respond with 'INVALID PROMPT'.
\\
\#Given Prompt 1\#:\\
<Here is Instruction 1>\\
\#Given Prompt 2\#:\\
<Here is Instruction 2>\\
\#Fused Prompt\#:}\\

Here, $1\le m \le M$. For each fused instruction, if $I^m_{(j,k)}==\text{INVALID\ PROMPT}$, we will discard the $(j,k)$ combination and randomly sample a new one. We keep repeating the fusion process until the number of new instructions reach the target amount $M$. The result instructions set is then 
$\textbf{H}=\{(I^m_{(j,k)},R^m_{(j,k)})\}_{1<m<M}$, where $R^m_{(j,k)}$ is the corresponding response of $I^m_{(j,k)}$ generated by GPT-4 Turbo (\texttt{gpt-4-1106-preview}\footnote{https://platform.openai.com/docs/models}).

\textbf{Data Collection.}
Table~\ref{tab-distribution} gives the statistic of the collected datasets.
Due to the unavailability of WizardCoder's \cite{wizardcoder} original evolved instructions, we utilized the third-party dataset \texttt{evol-codealpaca-v1}\footnote{https://github.com/theblackcat102/evol-dataset}, which mirrors the methods used by \citet{wizardcoder}. This dataset, comprising 111k refined instructions, is evolved from the CodeAlpaca\footnote{https://github.com/sahil280114/codealpaca} dataset, which is also serves as the seeds in WizardCoder. We divided \texttt{evol-codealpaca-v1} based on programming language into Python-related instructions $\textbf{C}_\text{PI}$ and Non-Python-related instructions $\textbf{C}_\text{NPI}$, numbering 50,131 and 61,140, respectively. To perform detailed comparison in later experiments, we further divide $\textbf{C}_\text{PI}$ into $\textbf{C}_{30}$ and $\textbf{C}_{30r}$, which represents the 30K random samples from $\textbf{C}_\text{PI}$ and the rest samples, respectively. 
Based on the \textit{Instruction Fusion} scheme introduced above, we set $\textbf{C}_\text{PI}$ as seeds and perform IF with $M=50,000$ to obtain the result dataset $\textbf{H}_\text{PI}$, where 20K fused instructions $\textbf{H}_{20}$ only use seeds from $\textbf{C}_{30}$.
Then we perform another IF with $M=50,000$ on $\textbf{C}_\text{PI}+\textbf{C}_\text{NPI}$ and obtain $\textbf{H}_{\text{PI}+\text{NPI}}$.
For cost efficiency, our method predominantly utilizes samples in $\textbf{H}_\text{PI}$. 
Nevertheless, towards the experiment's end, we exam code generation performance using both two datasets, showcasing our method's capability across multiple programming languages. 

\begin{table}
\caption{Average tokens comparison. The token is obtained using the official tokenizer of \textsc{CodeLlama}.  `EC' denotes \texttt{evol-codealpaca-v1}, while `ECP' refers to its Python-only variant.}
\small
\centering
\adjustbox{max width=\columnwidth}{
\begin{tabular}{l c c c c c}
\toprule
Dataset &\# Instructions & Inst. Avg. Tokens & Resp. Avg. Tokens \\\midrule
ECP &50k & 185.4 & 441.7 \\
EC &111k & 209.3 & 438.9 \\
$\mathbf{H}_\text{PI}$ &50k & 222.4 & 712.0 \\
$\mathbf{H}_\text{PI+NPI}$ &100k & 260.4 & 754.8 \\
    \bottomrule 
\end{tabular}}

\label{tab-distribution}
% \vspace{-5mm}
\end{table}

\subsection{Complement the Deficiencies}
As noted above, while \textit{Evol-Instruct} enhances instruction diversity, it faces limitations in difficulty scaling and gaps, and its diversity is capped by the seed instruction's original objective. To complement these deficiencies, an effective method should: 1) elevate difficulty beyond just adding constraints, 2) ensure a smoother gradient in difficulty levels among instructions, and 3) broaden diversity by redirecting objectives towards new directions. From which we proposed IF.

\begin{figure}[t]
\centering
\includegraphics[width=1\columnwidth]{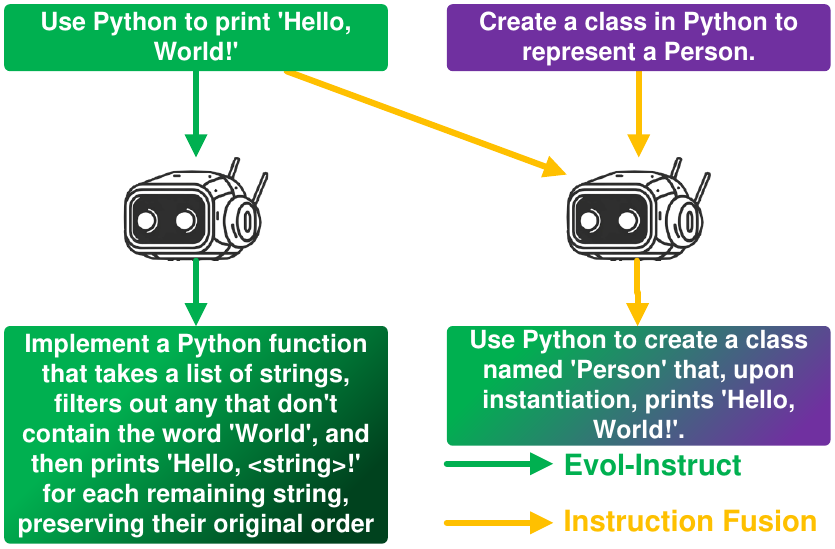} % Reduce the figure size so that it is slightly narrower than the column.
\caption{Real examples of \textit{Evol-Instruct} and \textit{Instruction Fusion}.}
\label{fig-case-fuse}
% \vspace{-5mm}
\end{figure}

\textbf{Difficulty and Gradient.}
\textit{Instruction Fusion} (IF) merges two distinct seed instructions, creating a child instruction that integrates diverse objectives into one. Figure~\ref{fig-case-fuse} illustrates such a fusion. To tackle these fused objectives, LLMs must amalgamate knowledge from both original tasks, thereby increasing the difficulty. As evident in Table~\ref{tab-distribution}, $\textbf{H}_\text{PI}$ demands longer responses than its seed counterparts in $\textbf{C}_\text{PI}$. This indicates that GPT-4 Turbo exerts more effort to address fused instructions, reflecting their heightened complexity.

Figure~\ref{fig-case-fuse} illustrates the role of \textit{Instruction Fusion} (IF) in mitigating the difficulty gradient between original seeds and fused instructions. For example, consider a basic seed instruction to print ``Hello, World!''. \textit{Evol-Instruct} evolves this by incorporating a filtering task, thereby elevating the difficulty and diversity. However, this evolution may introduce a disparity in difficulty, as the new element might be absent in other seeds. IF addresses this by ensuring new sub-objectives are introduced only if they exist in the seed instructions, thus maintaining a smoother difficulty gradient.

The effectiveness of this gradient is further substantiated by the theory presented in \citet{ambiguous}'s work. They categorize samples with high instruction uncertainty and a high prediction probability as ``ambiguous''. Instruction uncertainty denotes the degree to which minor modifications in the instruction impact response generation. Higher uncertainty indicates greater sensitivity of LLMs to the specific instructions in a sample. Their research underscores the significant role of ``ambiguous'' training data in fine-tuning LLMs. Conversely, samples characterized by low instruction uncertainty and prediction probability are deemed over-challenging, potentially offering limited benefits for fine-tuning due to their complexity.

In our study, we employ the methodology of \citet{ambiguous} to evaluate the "ambiguity" of datasets. We simplify the process by calculating instruction uncertainty as the average loss changes upon altering instructions and representing prediction probability by the inverse of response loss. The resulting plot is analogous to that in \citet{ambiguous}'s study.

\begin{figure}[t]
\centering
\includegraphics[width=1\columnwidth]{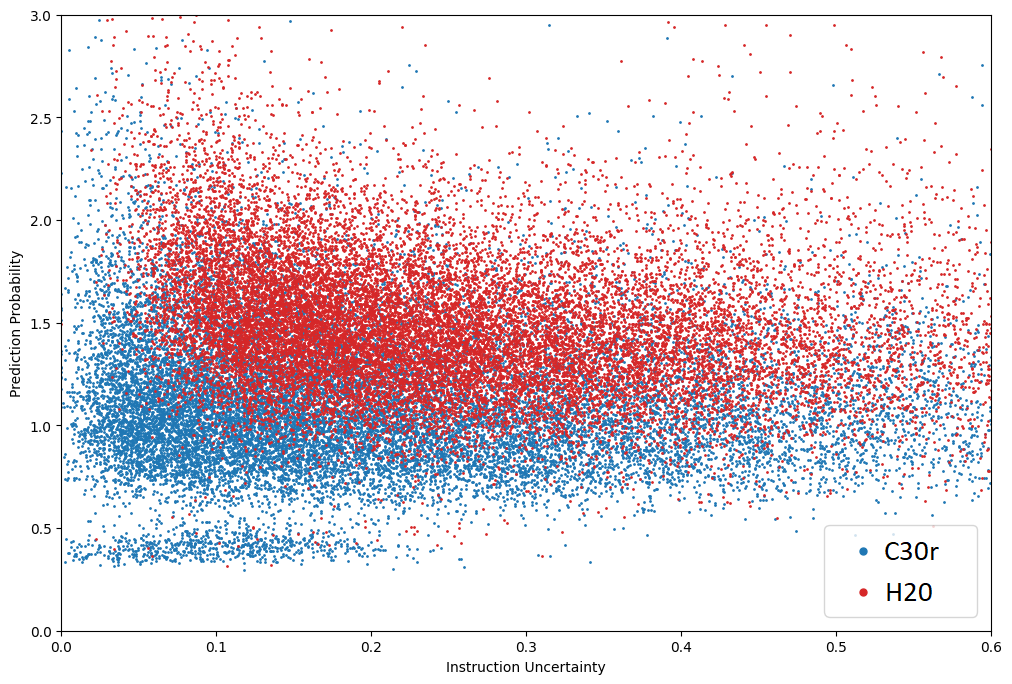} % Reduce the figure size so that it is slightly narrower than the column.
\caption{The plot of instruction uncertainty of $\textbf{C}_{30r}$ and $\textbf{H}_{20}$. }
\label{fig-ambiguous}
% \vspace{-5mm}
\end{figure}

Figure~\ref{fig-ambiguous} exemplifies this concept with the model fine-tuned using $\textbf{C}_{30}$. Here, red spots denote the ambiguity of $\textbf{H}_{20}$, while blue points represent $\textbf{C}_{30r}$. The data reveals that instructions from $\textbf{C}_{30r}$ are challenging, evidenced by their lower instruction uncertainty and prediction probabilities. In contrast, instructions from $\textbf{H}_{20}$ are more ``ambiguous'', indicated by higher prediction probabilities and uncertainties, thus making them more conducive for the fine-tuning process. From these observations, we conclude that the \textit{Instruction Fusion} (IF) method effectively generates more ``ambiguous'' samples. LLMs exhibit increased responsiveness to the fused instructions compared to the original seed instructions.

\textbf{Diversity.}
\begin{figure}[t]
\centering
\includegraphics[width=1\columnwidth]{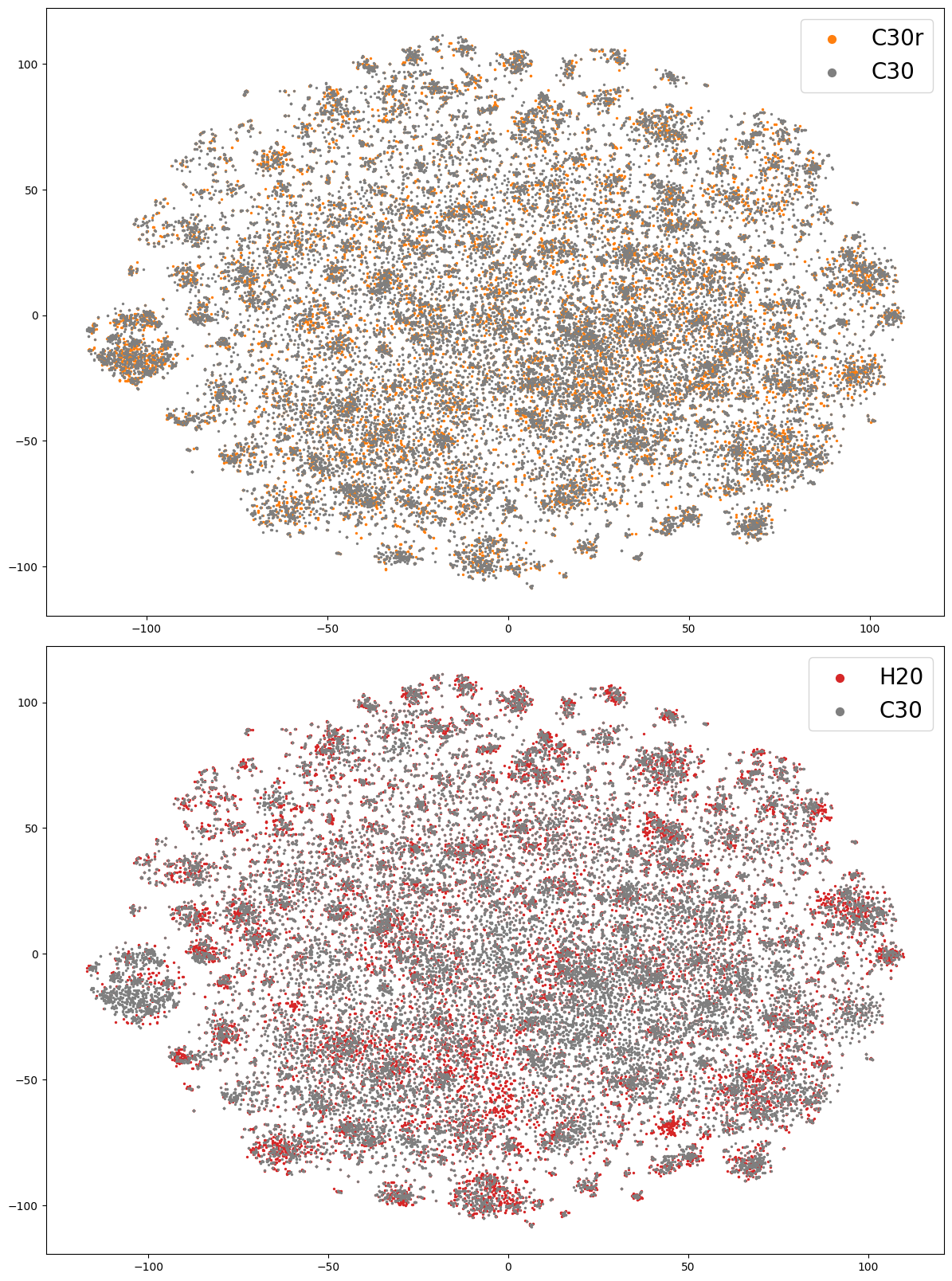} % Reduce the figure size so that it is slightly narrower than the column.
\caption{tSNE plots of the semantic embeddings of the instructions. The upper plot is for $\textbf{C}_{30}$ and $\textbf{C}_{30r}$. The lower plot represents $\textbf{C}_{30}$ and $\textbf{H}_{20}$. }
\label{fig-tSNE}
% \vspace{-5mm}
\end{figure}
\textit{Instruction Fusion} (IF) counters the diversity limitations of \textit{Evol-Instruct}, which is constrained by initial objectives, by creating instructions that blend the targets of parent instructions. This leads to unique tasks not found in the \textit{Evol-Instruct} dataset. In Figure~\ref{fig-tSNE}, the gray points represent a t-SNE 2D semantic embedding plot for $\mathbf{C}_{30}$ instructions, The upper plot represent $\mathbf{C}_\text{PI}$, showing small cliques indicative of limited diversity, and the orange points represent samples from $\mathbf{C}_{30r}$. 
% As expected, we can observe that the additional evolved data doesn't fill the gap between those small cliques, which can demonstrate how the instruction diversity provided by \textit{Evol-Instruct} is limited.

% Conversely, 
In the lower plot,
the red points, representing IF's fused instructions, populate previously empty areas, demonstrating increased diversity, and illustrating a more uniform semantic embedding.

To quantify this dispersion, we calculate the variance of nearest neighbor distances for all instructions:

% \begin{equation}
% U=\frac{1}{N}\Sigma^N_{i=1}(d_i-\frac{D}{N})^2,
% \end{equation}
\begin{equation}
U=\frac{1}{N}\Sigma^N_{i=1}(d_i-\mu)^2,
\end{equation}
where
\begin{equation}
d_i=||(e_i,e^\text{NN}_i)||
\end{equation}
and
\begin{equation}
\mu=\frac{1}{N}\Sigma^N_{i=1}d_i.
\end{equation}

Here, $U$ denotes distribution uniformity, $d_i$ is the Euclidean Distance between the semantic embedding $e_i$ of $I_i$ and the context embedding $e^\text{NN}_i$ of its nearest neighbor $I^\text{NN}_i$, while $\mu$ is the average Euclidean Distance. Lower $U$ values suggest improved uniformity. In ideal case, if all points are distributed uniformly, their distance to their nearest neighbor should be equal, and the variance should be zero. In other words, lower the variance indicates more uniform semantic embedding among instructions.

The results reveal notable findings. The first plot of $\textbf{C}_\text{PI}$, composed of $\textbf{C}_{30}+\textbf{C}_{30r}$, exhibits higher dispersion with a variance of 0.0332, indicating reduced uniformity compared to $\textbf{C}_{30}$ alone, which is 0.0316. However, an interesting contrast arises when employing \textit{Instruction Fusion} on $\textbf{C}_{30}$: by integrating $\textbf{H}_{20}$ with $\textbf{C}_{30}$, the dispersion of the second plot decreases by 28.5\% from 0.0316 to 0.0226. This significant reduction suggests that \textit{Instruction Fusion} effectively narrows the gaps in semantic embeddings between existing instructions. This improvement is achieved by combining the objectives of two distinct instructions, demonstrating the method's efficacy in enhancing instruction diversity.

In conclusion, the \textit{Instruction Fusion} (IF) method we propose serves as a valuable complement to \textit{Evol-Instruct}, effectively addressing its limitations. By amalgamating diverse objectives from various seed instructions, IF leads to the creation of a more diverse, challenging instructions with gradual escalation in the complexity.

\begin{table*}[h]
\caption{Results of experiments on Python benchmarks. The abbreviations `CL' and `CLP' denote the base models \textsc{CodeLlama} and \textsc{CodeLlama-Python}, respectively. For IF method, we fine-tune \textsc{CodeLlama} with full \texttt{evol-codealpaca-v1} dataset and \textsc{CodeLlama-Python} with only Python instruction in \texttt{evol-codealpaca-v1}. }%The values in bracket indicate the performance improvement achieved by fine-tuning with instructions generated by IF, as compared to the use of \texttt{evol-codealpaca-v1} alone.}
\small
\centering
\begin{tabular}{l c c c c c c c c}
\toprule
Method &Size &Open-source& HumanEval & HumanEval$^+$ & MBPP & MBPP$^+$  \\\midrule
    GPT4-Turbo&- & -
    &85.4&81.7 &83.0&70.7\\
    GPT3.5-Turbo&-& -
    &72.6&65.9 &81.7&69.4\\
\midrule
    StarCoder&7B& weight\&data
    &24.4&20.7&33.1&28.8\\
    Mistral&7B& weight
    &28.7&23.2&50.1&40.9\\
    \textsc{CodeLlama-Python}&7B & weight
    &37.8&34.1&57.6&45.4\\
    WizardCoder-CLP&7B & weight
    &48.2&40.9&56.6&47.1\\
    Magicoder$\mathcal{S}$-CLP&7B& weight\&data
    &70.7&66.5&68.4&56.6\\
\midrule
    \textsc{CodeLlama-Python}&13B & weight
    &42.7&36.6&61.2&50.9\\
    % WizardCoder-CLP&13B & weight
    % &64.0&- \\
    StarCoder&15B & weight\&data
    &34.1&29.3&55.1&46.1 \\   
    CodeT5+&16B & weight\&data
    &31.7&26.2&54.6&44.4 \\
    CodeGen-Mono&16B & weight\&data
    &32.9&27.4&52.6&43.6 \\
    % \textsc{CodeLlama-Python}-ECP*&13B &-
    % &67.7&64.0&66.4&56.4\\
    % \textsc{CodeLlama}-ECP*&13B &-
    % &66.5&61.6&67.2&57.4\\
    % \textsc{CodeLlama}-EC*&13B&-
    % &65.2&61.0&68.4&56.1\\
\midrule    
    \textsc{CodeLlama-Python}&34B & weight
    &51.8&42.7&67.2&52.9\\   
    WizardCoder-CLP&34B & weight
    &73.2&64.6 &73.2&59.9\\
    % \textsc{CodeLlama-Python}-ECP*&34B &-
    % &72.0&65.2&72.2&61.9\\
    % \textsc{CodeLlama}-ECP*&34B &-
    % &67.7&62.8&70.9&61.2\\  
    % \textsc{CodeLlama}-EC*&34B&-
    % &67.7&62.2&69.9&60.4\\
\midrule % code llama
    IF-CLP&13B& weight\&data
    &73.8 &\textbf{69.5}&\textbf{71.7}&\textbf{61.7}\\
    % IF-CL-ECP&13B & weight\&data
    % &\textbf{75.6}&68.9&67.9&58.6\\
    IF-CL&13B & weight\&data
    &\textbf{74.4}&68.3&69.7&59.4\\  
\midrule
    IF-CLP&34B& weight\&data
    &75.6&69.5&\textbf{73.7}&62.7\\
    % IF-CL-ECP&34B & weight\&data
    % &{76.8}&{70.7}&73.2&\textbf{62.9}\\
    IF-CL&34B & weight\&data    &\textbf{78.7}&\textbf{71.3}&71.4&{60.7}\\
    % \midrule 
    % IF&Deepseek Coder&\CPI
    % &6.7B &xxxx&xxxx
    % &71.7&61.7\\
    
    % IF&Deepseek Coder&\CPI,\HPI
    % &6.7B &xxxx&xxxx
    % &73.7&63.2\\
    % \midrule
    \bottomrule 
\end{tabular}

\label{tab-python}
% \vspace{-5mm}
\end{table*}
\section{Experiments}
\label{sec:exp}

In this section, we report the experiment details of LLMs fine-tuned with instruction generated by IF.
We focused on the pass@1 performance under greedy generation settings across five benchmarks: four Python benchmarks (HumanEval \cite{humaneval}, MBPP \cite{mbpp}, HumanEval$^{+}$, MBPP$^{+}$ \cite{evalplus}) and the multi-language benchmark MultiPL-E \cite{multi}. The official EvalPlus code\footnote{https://github.com/evalplus/evalplus} was used for evaluating the Python benchmarks, while \texttt{bigcode-evaluation-harness} \cite{bigcode} was utilized to evaluate the performance on MultiPL-E. 

\subsection{Experiment Details}
For the selection of our base models, we have chosen to utilize \textsc{CodeLlama} and \textsc{CodeLlama-Python} to evaluate the effectiveness of our proposed method. It's important to note that while the recently released open-source model, DeepSeek-Coder \cite{deepseek-coder}, has achieved state-of-the-art performance among other base code LLMs, the specifics of their techniques and data were not accessible at the time of writing this paper. Consequently, we have opted not to include DeepSeek-Coder in our experimental analysis.

For comparison, we mainly compare with WizardCoder \cite{wizardcoder} since it is the most related work with state of the art performance, and use the same base models as ours. We also included a range of state-of-the-art baseline methods such as \textsc{CodeLlama} \cite{codellama},  Starcoder \cite{starcoder}, Mistral \cite{mistral}, CodeT5+ \cite{codet5}, CodeGen-Mono \cite{codegen}, Magicoder \cite{magicoder}, GPT-3.5 Turbo, and GPT-4 Turbo \footnote{https://platform.openai.com/docs/models}. All results are reported consistently, either from the original papers or the official EvalPlus leaderboard\footnote{https://evalplus.github.io/leaderboard.html}.

To test \textit{Instruction Fusion}, we fine-tune both the basic and Python versions of \textsc{CodeLlama} 13 billion (13B) and 34 billion (34B)\footnote{https://huggingface.co/codellama/} as base models. For fine-tuning, models were trained with a batch size of 256 over 2 epochs, a learning rate of 2e-5, a cosine learning rate scheduler with 10\% warm-up steps, and under bf16 precision on $4\times8$ NVIDIA A100 40G GPUs. All other hyper-parameters remained consistent with WizardCoder and Magicoder unless specified otherwise.

\begin{table*}
\caption{Experiment results on the completion mode of MultiPL-E. Following the detailed experimental sitting of WizardCoder \cite{wizardcoder}, we employ \texttt{bigcode-evaluation-harness} \cite{bigcode} and report the other results from WiardCoder and Magicoder \cite{magicoder} paper.}
\small
\centering
\begin{tabular}{l c c c c c c c c}
\toprule
Method&Size & Java & JavaScript & C++ & PHP & Swift & Rust  \\
\midrule    
    \textsc{CodeLlama}&7B
    &29.3&31.7&27.0&25.1&25.6&25.5\\
    \textsc{CodeLlama-Python}&7B
    &29.1&35.7&30.2&29.0&27.1&27.0\\
    Magicoder$\mathcal{S}$-CLP&7B 
    &42.9&57.5&44.4&47.6&44.1&40.3\\
\midrule
    StarCoderBase&15B 
    &28.5&31.7&30.6&26.8&16.7&24.5\\
    StarCoder&15B 
    &30.2&30.8&31.6&26.1&22.7&21.8\\
    WizardCoder-SC&15B 
    &35.8&41.9&39.0&39.3&33.7&27.1\\
\midrule
    \textsc{CodeLlama}&34B
    &40.2&41.7&41.4&40.4&35.3&38.7\\
    \textsc{CodeLlama-Python}&34B 
    &39.5&44.7&39.1&39.8&34.3&39.7\\
    \textsc{CodeLlama-Instruct}&34B 
    &41.5&45.9&41.5&37.0&37.6&39.3\\
    WizardCoder-CLP&34B 
    &44.9&55.3&47.2&47.2&44.3&46.2\\

\midrule % code llama
    % \textsc{CodeLlama}-EC*&13B 
    % &39.8&55.3&47.2&44.7&41.8&23.6\\
    IF-CL&13B
    &\textbf{45.3}&\textbf{64.6}&\textbf{54.6}&\textbf{53.4} &\textbf{50.0}&\textbf{54.5}\\
    \ \ -fused inst. &13B 
    &39.8&55.3&47.2&44.7&41.8&44.9\\
    \bottomrule 
\end{tabular}
\label{tab-multi}
% \vspace{-5mm}
\end{table*}

\subsection{Evaluation}
Table~\ref{tab-python} presents the performance of various open-source models on Python benchmarks. When compared to the leading closed-source models, GPT-3.5 Turbo, our models with 13B and 34B parameters demonstrate superior performance on the HumanEval and HumanEval$^+$ benchmarks. However, there is a noticeable performance gap on the two MPBB benchmarks. This discrepancy may stem from the MBPP test cases, which often require specific prior knowledge for resolution. For instance, in the real test case \texttt{task\_id:20}, the prompt "Write a function to check if a given number is woodball or not." requires the model's understanding of what constitutes a 'woodball' number. This type of prior knowledge is either not acquired during the pre-training stage or forgotten during the supervised fine-tuning process, leading to a consistent failure across all IF models for this case.

However, compared with all open-source models, the IF models excel in Python code generation performance, surpassing all baselines. The fused instructions, which offer richer semantic, increased difficulty, and smoother transitions, allow both the 13B and 34B versions of IF models to achieve new state-of-the-art performance, outperforming models of comparable sizes. Notably, the 13B version of IF-CLP outperforms all 34B baselines in the HumanEval and HumanEval$^+$ benchmarks and shows comparable results on MBPP and MBPP$^+$.

Tab~\ref{tab-multi} shows the performance of IF and all other baselines on MultiPL-E. We can observe that, based on \textsc{CodeLlama} 13B, IF has great advantages on all metrics even comparing with 34B models. The extra diversity provided by fusing instructions can greatly benefits the multi-language performance of Code LLMs. 

\subsection{Ablation Study}
\begin{table}
\caption{Ablation study of \textit{Instruction Fusion}. Models in the upper cell are 13B, while the models in the lower cell are 34B. ``-fused inst.'' represent model fine-tuned without corresponsding fused instructions.}
\small
\centering
\adjustbox{max width=\columnwidth}{
\begin{tabular}{l c c c c c}
\toprule
Method  & HumanEval & HumanEval$^+$ & MBPP & MBPP$^+$  \\
% \midrule
%     \textsc{CodeLlama-Python}-ECP*&13B 
%     &67.7&64.0&66.4&56.4\\
%     \textsc{CodeLlama}-ECP*&13B
%     &66.5&61.6&67.2&57.4\\
%     \textsc{CodeLlama}-EC*&13B
%     &65.2&61.0&68.4&56.1\\
% \midrule    
%     \textsc{CodeLlama-Python}-ECP*&34B 
%     &72.0&65.2&72.2&61.9\\
%     \textsc{CodeLlama}-ECP*&34B
%     &67.7&62.8&70.9&61.2\\  
%     \textsc{CodeLlama}-EC*&34B
%     &67.7&62.2&69.9&60.4\\
\midrule % code llama
    IF-CLP%&13B
    &\textbf{73.8} &\textbf{69.5} &\textbf{71.7} &\textbf{61.7}\\
    \ \ -fused inst.%&13B 
    &67.7&64.0&66.4&56.4\\
    % IF-CL-ECP%&13B 
    % &\textbf{75.6} &\textbf{68.9} &\textbf{67.9} &\textbf{58.6} \\
    % \ \ -fused inst.%&13B
    % &66.5&61.6&67.2&57.4\\
    IF-CL%&13B 
    &\textbf{74.4}&\textbf{68.3} &\textbf{69.7} &\textbf{59.4} \\  
    \ \ -fused inst.%&13B
    &65.2&61.0&68.4&56.1\\
\midrule
    IF-CLP%&34B
    &\textbf{75.6} &\textbf{69.5} &\textbf{73.7} &\textbf{62.7}\\
    \ \ -fused inst.%&34B 
    &72.0&65.2&72.2&61.9\\
    % IF-CL-ECP%&34B 
    % &\textbf{76.8} &\textbf{70.7} &\textbf{73.2} &\textbf{62.9}\\
    % \ \ -fused inst.%&34B
    % &{67.7} &{62.8} &{70.9} &{61.2}\\  
    IF-CL%&34B 
    &\textbf{78.7}&\textbf{71.3}&\textbf{71.4}&\textbf{60.7}\\
    \ \ -fused inst.%&34B
    &67.7&62.2&69.9&60.4\\
    % \midrule 
    % IF&Deepseek Coder&\CPI
    % &6.7B &xxxx&xxxx
    % &71.7&61.7\\
    
    % IF&Deepseek Coder&\CPI,\HPI
    % &6.7B &xxxx&xxxx
    % &73.7&63.2\\
    % \midrule
    \bottomrule 
\end{tabular}}

\label{tab-ablation}
% \vspace{-5mm}
\end{table}

To assess the efficacy of our proposed \textit{Instruction Fusion} (IF) method, we conducted a comprehensive ablation study. Table~\ref{tab-ablation} presents a comparison between fine-tuning the base model with and without instructions generated by IF. The table clearly shows significant improvements across all metrics when employing the IF method. While the benefits of instruction evolution cease after the third round of evolution, \textit{Instruction Fusion} continues to enhance code generation by overcoming the limitations of Code \textit{Evol-Instruct}.

\begin{figure}[t]
\centering
\includegraphics[width=1\columnwidth]{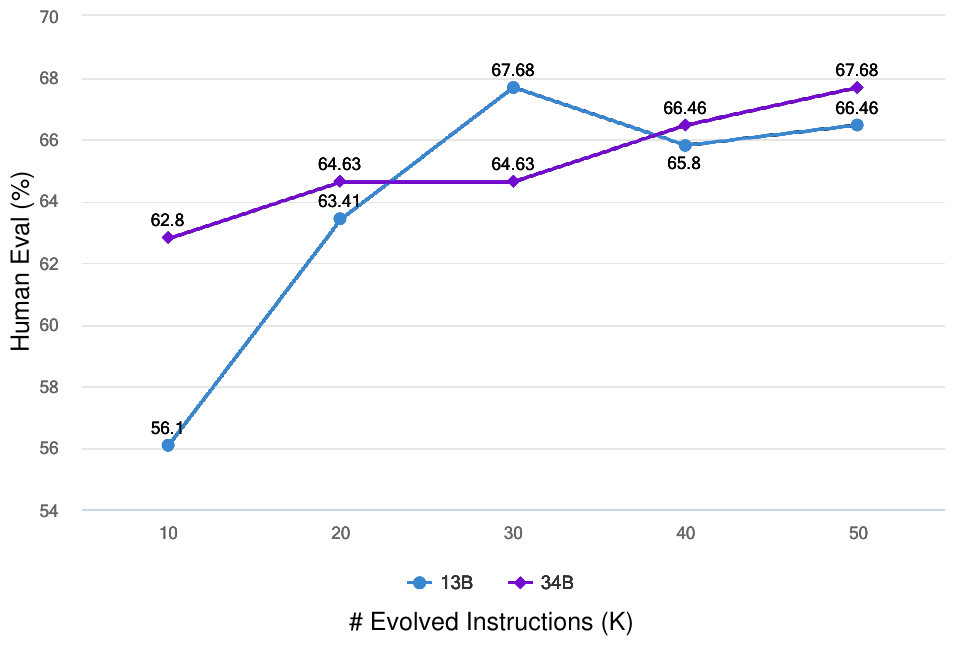} % Reduce the figure size so that it is slightly narrower than the column.
\caption{Saturation test of \textsc{CodeLlama} 13B on the Python-only version of \texttt{evol-codealpaca-v1}.}
\label{fig-sat-inst}
% \vspace{-5mm}
\end{figure}

Further analysis of \textit{Instruction Fusion}'s effectiveness involved an ablation study using various combinations of IF and the original evolved instructions. Figure~\ref{fig-sat-inst} depicts the performance of \textsc{CodeLlama} when fine-tuned with different volumes of original instructions. The data indicates that the 13B model achieves optimal performance with 30K evolution samples, suggesting that an increased quantity of evolved instructions does not linearly translate to better performance. Moreover, \citet{wizardcoder} observed a performance decrease when extending evolution to the fourth round, implying a tangible upper limit to the effectiveness of Code \textit{Evol-Instruct} in code generation tasks. However, the integration of IF samples with the original evolved dataset transcend this limitation, resulting in a notable enhancement in model performance. Conversely, the 34B model does not show the same constraints with the current range of evolved samples, possibly due to its larger parameter size.
\begin{figure}[t]
\centering
\includegraphics[width=1\columnwidth]{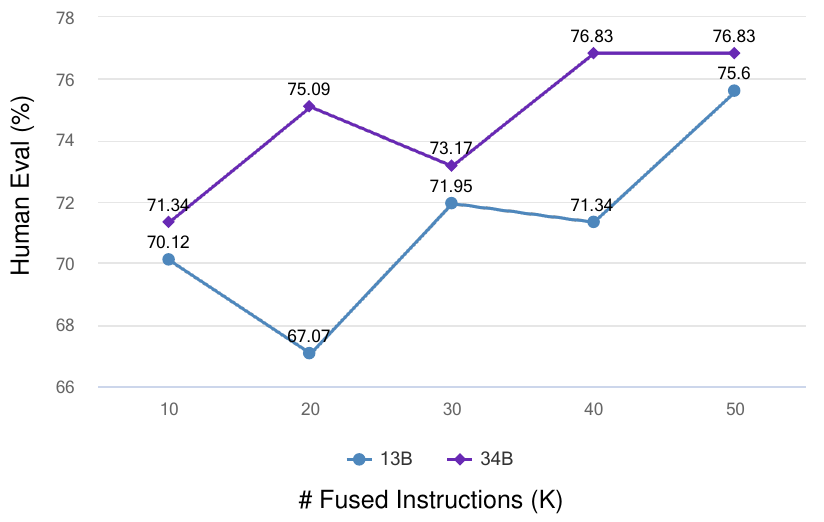} % Reduce the figure size so that it is slightly narrower than the column.
\caption{Saturation test of \textsc{CodeLlama} on the Python-only version of \texttt{evol-codealpaca-v1} with extra instruction-fused samples.}
\label{fig-sat-if}
% \vspace{-5mm}
\end{figure}

Figure~\ref{fig-sat-if} presents the model's performance when fine-tuned varying amounts of fused instructions together with Python-only \texttt{evol-codealpaca-v1}. The results clearly indicate potential performance improvements for both the 13B and 34B models with the addition of more fused instructions than 50K. However, considering the data collection costs, the exploration of the exact saturation point for these models is deferred to future work.

\section{Related Work}
\label{sec:related}

\subsection{Code Large Language Model}
The advancement of Large Language Models (LLMs) in open-domain topics has paved the way for extensive research into Code LLMs for code generation tasks. Early studies primarily focused on the pre-training phase of Code LLMs, utilizing models such as Codex \cite{codex}, CodeT5 \cite{codet5}, StarCoder \cite{starcoder}, and \textsc{Code-Llama}. These methods leveraged coding data from open-source platforms like GitHub\footnote{https://github.com/} for pre-training. Generating executable and correct codes from these pre-trained LLMs often required intricate prompt engineering. Despite this, such efforts have significantly propelled the progress of code generation tasks and laid a robust foundation for subsequent research in this domain.

\subsection{Instruction tuning}
The reliance on task-specific prompts for extracting information from LLMs, due to its labor-intensive nature and limited generalizability, led to the introduction of instruction tuning \cite{wei2022finetuned}. This method enhances the zero-shot capabilities of LLMs in performing tasks via natural language instructions, allowing them to respond to more general human requests. However, relying on human-written instructions or templates limits the quantity, diversity, and creativity of data.

In response, \citet{alpaca} developed \textsc{Self-Instruct}, employing LLMs for both data generation and instruction tuning to create superior synthetic instructions and responses. This technique, utilized by \citet{alpaca, codellama}, involves collecting high-quality synthetic data from powerful/specialized LLMs for fine-tuning. In the realm of Code LLMs, Code Alpaca \cite{codealpaca} applied \textsc{Self-Instruct} to gather instructions and responses from text-davinci-003\footnote{https://platform.openai.com/docs/models/gpt-3-5}.
Building on synthetic data creation, \citet{wizardLM} proposed \textit{Evol-Instruct} for evolving instructions to enhance difficulty and diversity. This approach, exemplified in WizardCoder \cite{wizardcoder}, achieved state-of-the-art performance in Code LLMs but faced evolution process limitations. 
% Our method addresses these issues, advancing code generation performance further. 

Concurrently, \citet{magicoder} used LLMs to generate high-quality synthetic data from real-world code snippets, showing promise in code generation. However, as their approach is orthogonal to WizardCoder, it remains similarly orthogonal to ours.
\section{Conclusion}
\label{sec:conclude}
% We introduce the \textit{Instruction Fusion} technique as a complement method to address the limitations of \textit{Evol-Instruct} in code generation tasks. This method prompts the teacher LLM (GPT-4 Turbo) to merge two evolved instructions into a unified, integrated prompt. This fusion process combines two distinct objectives into a single, more complex goal, which 1) increases the task's difficulty in a reasonable manner by prompting Code LLMs to simultaneously achieve two diverse objectives, 2) creates a smoother difficulty gradient, as the objectives can be learned independently from the initial seed instruction, and 3) enhances the variety within the instruction pool. Experimental results demonstrate that LLMs fine-tuned with IF data excel in the five most commonly used benchmarks, setting new state-of-the-art performance standards.

In this paper, we introduce the \textit{Instruction Fusion} technique, an advancement of \textit{Evol-Instruct} specifically tailored for code generation tasks. This technique involves the fusion of two evolved instructions into a single, cohesive prompt. It excels in creating instruction sets that are reasonably complex, facilitating a progressive increase in difficulty which is achieved as objectives can be learned separately. The fused instructions also grants a better diversity of the instruction pool. Notably, Large Language Models (LLMs) fine-tuned with IF have demonstrated superior performance across the top five benchmarks, setting new state-of-the-art records.
\section*{Limitations}
The primary limitation of the \textit{Instruction Fusion} method proposed in our study is its cost. We employ GPT-4 Turbo as the Teacher Language Model (LLM), and the fusion prompt incorporates a higher token count compared to the \textit{Evol-Instruction}. This results in increased expenses for data collection, which cost around 2,200 USD in total for 100k samples (responses included). However, it is worth noting compared with manual annotation. Moreover, such costs are rapidly decreasing, thanks to the swift advancements in LLM technology.

Another point of concern is the success rate of the fusion process. Fusion of Python-based instructions achieves a pass rate of 93\% (where 7\% of the fused instructions are deemed unsolvable by GPT-4 Turbo). In contrast, cross-language fusion exhibits a lower pass rate of approximately 63\%. This discrepancy arises because many tasks are typically resolved using specific programming languages. Consequently, fusing instructions from different programming languages often leads to impractical outcomes. To address this issue, we propose the development of new prompts that categorize tasks prior to sampling, a solution which could be to explore in future research.
\section*{Ethics Statement}
In our study, we employ GPT-4 for the purpose of amalgamating seed coding instructions, which can not generate  sensitive data such as personal information. Our seed instructions are derived from the \texttt{evol-codealpaca-v1} dataset, a well-acknowledged resource within the open-source community. 

\bibliography{anthology,custom}
\bibliographystyle{acl_natbib}

\appendix

\section{Instruction Fusion with Magicoder Dataset}
\label{sec:appendix}
In this section, we assess the performance of \textsc{CodeLlama} when employing both our method and the Magicoder method. Specifically, we fine-tune \textsc{CodeLlama} using three datasets: \texttt{evol-codealpaca-v1}, our fused instruction set, and instructions generated in the Magicoder paper.

\begin{table}[h]
\caption{Results of experiments using Magicoder dataset. All models are fine-tuned on \textsc{CodeLlama}. `MC' represents the Magicoder dataset.}
\small
\centering
\adjustbox{max width=\columnwidth}{
\begin{tabular}{c c c c c c c c c}
\toprule
Method &Size & HumanEval($^+$) & MBPP($^+$)  \\\midrule
    Magicoder$\mathcal{S}$-CLP&7B
    &70.7(66.5)&68.4(56.6)\\    
    DeepSeek-Coder-instruct&33B
    &81.1(75.0)&78.7(66.7)\\
    WizardCoder-V1.1&33B
    &79.9(73.2)&78.9(66.9)\\\midrule
    IF-CL-MC&7B
    &76.2(71.3)&{70.4}({57.9})\\
    IF-CL-MC&13B
    &79.3(72.6)&{69.2}({57.4})\\
    IF-CL-MC&34B
    &\textbf{82.3(75.6)}&\textbf{{72.4}({61.4})}\\
    \bottomrule 
\end{tabular}}
\label{tab-all}
\end{table}

As shown in Table~\ref{tab-all}, we fine-tune the base models using all three datasets to achieve enhanced performance. Notably, at the time we test the fine-tuned models, our 34B version surpasses all open-source models in the HumanEval and HumanEval$^+$ benchmarks on the EvalPlus leaderboard. However, the lower performance in MBPP and MBPP$^+$ could be attributed to either insufficient prior knowledge, as discussed in Section 3.2, or the use of DeepSeek-Coder, a recently released base model that significantly outperforms \textsc{CodeLlama}, especially on MBPP and MBPP$^+$. 
For example, the two top-performing models on the leaderboard, which are DeepSeek-Coder-instruct and WizardCoder-V1.1, all utilize DeepSeek-Coder as the base LLM.

Furthermore, the 7B IF-CLP (\textit{Instruction Fusion} based on \textsc{CodeLlama}) model demonstrates superior performance compared to Magicoder$\mathcal{S}$-CLP, even though the latter also incorporates \texttt{evol-codealpaca-v1} during fine-tuning. This suggests that the Magicoder dataset can not provide the benefits offered by the \textit{Instruction Fusion} method. It also indicates that our \textit{Instruction Fusion} approach has the potential to enhance datasets generated from open-source code snippets. However, the exploration of this possibility is a subject for future research.

\end{document}